\documentclass[runningheads]{llncs}
\usepackage[T1]{fontenc}

\usepackage{graphicx,verbatim}
\usepackage{xcolor}
\usepackage{amssymb}
\usepackage{amsmath,amssymb,amsfonts}%

\usepackage{hyperref}

\begin{document}
\title{Towards MR-Based Trochleoplasty Planning}

\author{Michael Wehrli\inst{1}\orcidID{0009-0005-8740-3295} \and
Alicia Durrer \inst{1}\orcidID{0009-0007-8970-909X} \and
Paul Friedrich \inst{1}\orcidID{0000-0003-3653-5624} \and
Sidaty El Hadramy \inst{1}\orcidID{0009-0000-2917-0706} \and
Edwin Li \inst{2}\orcidID{0000-0001-9881-5344} \and
Luana Brahaj \inst{1} \and
Carol C. Hasler \inst{3} \and
Philippe C. Cattin \inst{1}\orcidID{0000-0001-8785-2713}
}
%
\authorrunning{M. Wehrli et al.}

\institute{Department of Biomedical Engineering, University of Basel, Allschwil, Switzerland \and
Olten Cantonal Hospital, Olten, Switzerland \and
Department of Orthopedics, University Children's Hospital Basel, Basel, Switzerland
}
    
\maketitle              

\begin{abstract}
To treat Trochlear Dysplasia (TD), current approaches rely mainly on low-resolution clinical Magnetic Resonance (MR) scans and surgical intuition. The surgeries are planned based on surgeons experience, have limited adoption of minimally invasive techniques, and lead to inconsistent outcomes. We propose a pipeline that generates super-resolved, patient-specific 3D pseudo-healthy target morphologies from conventional clinical MR scans. First, we compute an isotropic super-resolved MR volume using an Implicit Neural Representation (INR). Next, we segment femur, tibia, patella, and fibula with a multi-label custom-trained network. Finally, we train a Wavelet Diffusion Model (WDM) to generate pseudo-healthy target morphologies of the trochlear region. In contrast to prior work producing pseudo-healthy low-resolution 3D MR images, our approach enables the generation of sub-millimeter resolved 3D shapes compatible for pre- and intraoperative use. These can serve as preoperative blueprints for reshaping the femoral groove while preserving the native patella articulation. Furthermore, and in contrast to other work, we do not require a CT for our pipeline - reducing the amount of radiation. We evaluated our approach on 25 TD patients and could show that our target morphologies significantly improve the sulcus angle (SA) and trochlear groove depth (TGD). The code and interactive visualization are available at \url{https://wehrlimi.github.io/sr-3d-planning/}.


\end{abstract}
\section{Introduction}
\begin{figure}[h]
    \centering
    \includegraphics[width=0.9\textwidth]{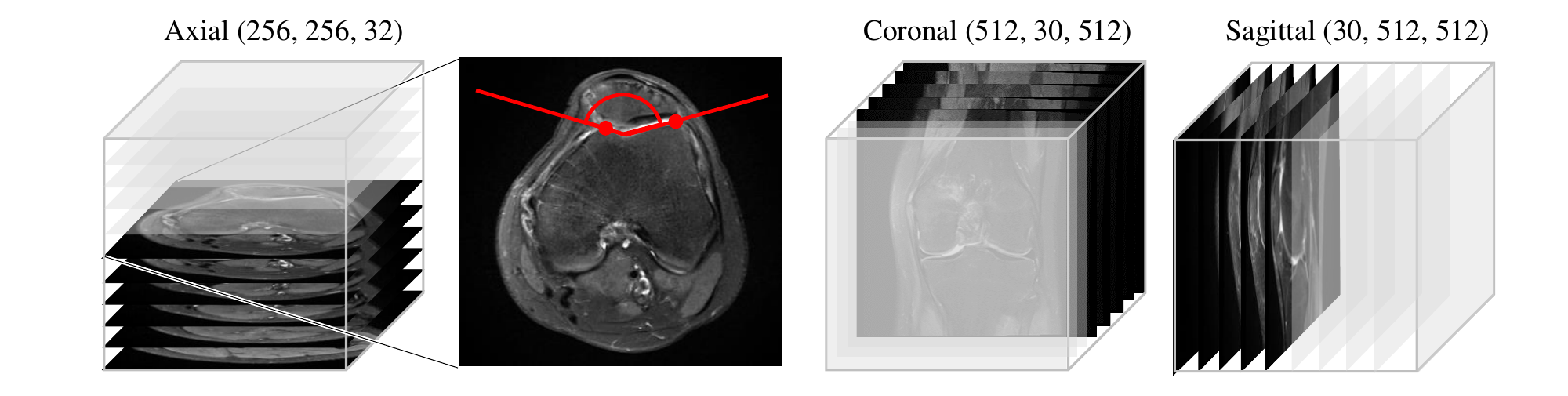}
    \caption{Currently, the diagnosis of Trochlear Dysplasia (TD) relies on three orthogonal MR volumes: axial, sagittal, and coronal. TD shows an abnormally large sulcus angle (visualized in red)~\label{fig_Intro} that can be seen in the axial volume. Limited resolution and inter-slice spacing impair accurate measurements.}
    \label{fig:volumes}
\end{figure}

Trochlear Dysplasia (TD) is an anatomical deformity of the femoral trochlea, often seen in adolescents with anterior knee pain and patellar instability~\cite{batailler2018trochlear}. The abnormal trochlear shape compromises patellar tracking, increasing the risk of dislocation and long-term degenerative changes, such as osteoarthritis~\cite{hasler2016patella}. While trochleoplasty is a recognized surgical procedure for reshaping the dysplastic trochlear groove in individuals with patellar instability, numerous patients still experience ongoing pain and exhibit varying levels of satisfaction~\cite{blond2023trochlear}. This is largely due to the procedure's complexity, mainly qualitative planning and limited amount of discrete measurements to describe TD~\cite{van2014statistical}. In current practice, TD diagnosis and surgical planning are based on MR Imaging in three orthogonal planes (axial, sagittal, coronal), each with limited resolution and spacing~\cite{nacey2017magnetic}, see \autoref{fig_Intro}. Radiologists manually assess the sulcus angle and classify TD using the Déjour criterion~\cite{batailler2018trochlear,beaufils2012trochleoplasty}. However, intraoperative decisions are guided by the surgeon's clinical experience rather than accurate, patient-specific preoperative planning. The surgeon must reshape the femur to fit the individual's patella, a task complicated by the anatomical variability. Without consistent planning to evaluate the deformity and guide surgical decisions, achieving optimal outcomes and reproducibility becomes significantly more challenging and only experts can perform arthroscopic trochleoplasty~\cite{blond2015arthroscopic}.

A useful preoperative plan should be based on 3D high-resolution images that allow accurate measurements and the definition of a surgical target morphology for intraoperative navigation~\cite{fang2024patient}. Multiple studies have attempted to generate preoperative guidance for TD~\cite{cerveri2018stacked,cerveri2019representative,lee2024shallow,van2014statistical}. But all of these are based on CT scans that expose patients to harmful radiation. As trochleoplasty is often performed in adolescents whose anatomy is not yet fully developed, high-resolution, low-risk imaging is essential. A recent study~\cite{wehrli2025generating} investigated the generation of pseudo-healthy images from MR scans using wavelet diffusion models. While this approach avoids radiation, it is limited to inpainting axial volumes, shown in the left part of \autoref{fig_Intro}. It can be used to give the surgeon an idea of the pseudo-healthy shape. However, it cannot support intraoperative planning, as the generated images still require segmentation, a step that is unreliable on data that has synthetic inpaintings. Segmenting synthetic images remains problematic, as they often contain out of distribution features. Popular segmentation models~\cite{d2025totalsegmentator,gu2025segmentanybone}, trained on real data, fail to generalize to these. To overcome these limitations, we propose a novel MR-only pipeline, visualized in \autoref{fig:Fig_Method}. We generate a patient-specific high-resolution 3D multi-label segmentation that reflects the expected healthy anatomy and can directly support intraoperative guidance. Our contributions are: 

\begin{itemize}
    \item A super-resolution approach leveraging implicit neural representations (INRs) to create isotropic, high-detail knee MR volumes from conventional clinical knee MR scans.
    \item A segment-first pipeline for pseudo-healthy 3D planning, avoiding segmentation of artificially inpainted images.
    \item To the best of our knowledge, our method is the first to generate surgical target morphologies for TD from MR Imaging alone. It bridges the diagnostic and surgical planning gap, promoting safer, more consistent, and potentially less invasive trochleoplasty.
\end{itemize}
\section{Methods}
\begin{figure}[h]
    \centering
    \includegraphics[width=1\linewidth]{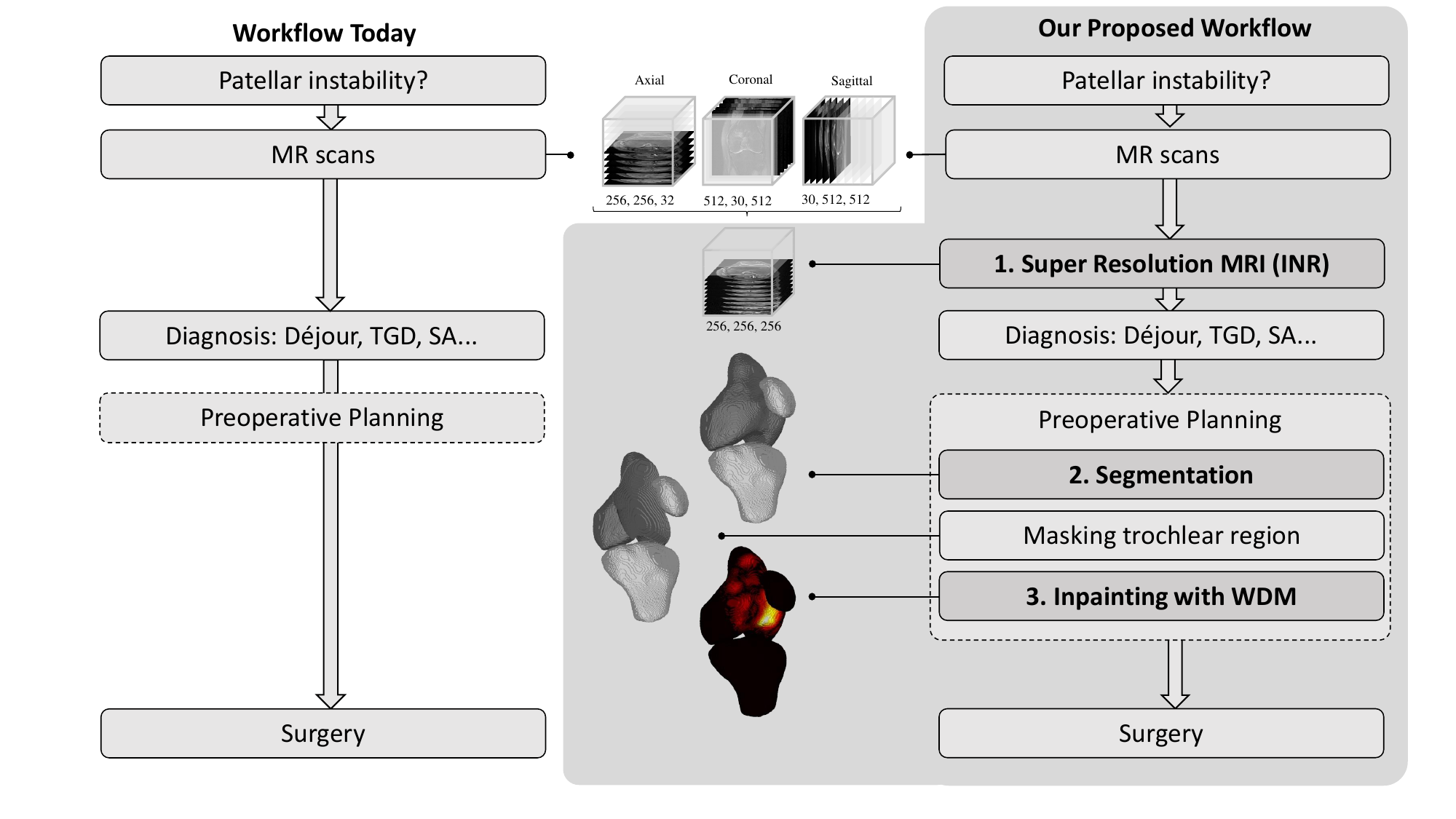}
    \caption{Left: The workflow today. Trochleoplasty is commonly performed without preoperative planning, relying on surgical expertise. Right: Our proposed workflow enables patient-specific super-resolved 3D planning, visualized as a 3D heatmap between the patients original knee with TD and it's pseudo-healthy version.}
    \label{fig:Fig_Method}
\end{figure}

We propose a novel MR-only workflow for the generation of patient-specific 3D target morphologies in TD, illustrated in \autoref{fig:Fig_Method}. Our method proceeds in three key steps:
\begin{enumerate}
    \item We reconstruct isotropic super-resolved 3D MR volumes from clinically acquired scans using INRs.
    \item We manually annotated 40 super-resolved volumes. We trained a segmentation model with those labels to extract key anatomical structures.
    \item We apply a Wavelet Diffusion Model (WDM) to inpaint the pathological region, generating a pseudo-healthy 3D volume.
\end{enumerate}
The resulting inpainted multi-label segmentations can be converted into 3D meshes for intraoperative navigation. Each step is described in more detail below.

\subsection{Step 1: Super Resolution MR Imaging (INR)}
\label{subsec:INR}
Clinical knee MR images are often anisotropic due to significant inter-slice spacing resulting from acquisition along different axes with high in-plane and low out-of-plane resolution. Thus, our goal is to fuse different scans into a single, high-resolution, isotropic volume. Given 3 MR images acquired along different axes (as shown in \autoref{fig:volumes}) and potentially different contrasts, we aim to use implicit neural representations (INRs) to fuse all available information by learning a shared representation of the underlying anatomy. Building upon \cite{mcginnis2023single}, we define a continuous function $f_{\theta}:\mathbb{R}^C\rightarrow\mathbb{R}^D$ that maps a spatial coordinate $c\in\mathbb{R}^{C}$ to the image intensity values $d\in\mathbb{R}^{D}$. By jointly optimizing this function to represent all 3 scans, it implicitly learns the shared anatomical features present in the different scans, resembling a unified representation. We define this function as a $5$-layer MLP with WIRE activations \cite{saragadam2023wire}, a hidden dimension of $1024$, and $3$ output channels (one per scan), and jointly optimize it, by minimizing the pixel-wise mean-squared error (MSE) between the model's prediction, and a given value within a set of known pixel-value pairs $\mathcal{C}$:
\begin{equation}
    \mathcal{L}_{MSE} = \sum_{c\in \mathcal{C}}\| f_\theta(c) - d(c) \|_2^2 .
\end{equation}
Each model is optimized for $100$ epochs, with a batch size of $4096$ pixel-value pairs per iteration. We use an Adam optimizer with a learning rate of $4\times10^{-4}$ that follows a cosine annealing learning rate schedule. To obtain a super-resolved volume, we evaluate the learned function over a given spatial domain, using an arbitrary resolution (in our case $256^3$).

\subsection{Step 2: Segmentation}
\label{subsec:SEG}
We annotated 10 super-resolved MR volumes from our in-house dataset and 30 super-resolved MR volumes from the fastMRI dataset, see Section~\ref{sec:data}. Four anatomical structures were labeled: Femur, tibia, patella and fibula. We used 3D Slicer~\cite{slicer,kikinis20133d} with the MONAI Label framework~\cite{diaz2024monai}. The manual segmentations were reviewed by a deputy attending orthopedic surgeon. We chose a 3D U-Net (with a pretrained SegResNet~\cite{myronenko20183d}) using a Dice and cross-entropy loss.

\subsection{Step 3: Inpainting with WDM}
\label{subsec:WDM}
\begin{figure}[h]
    \centering
    \includegraphics[width=0.9\linewidth]{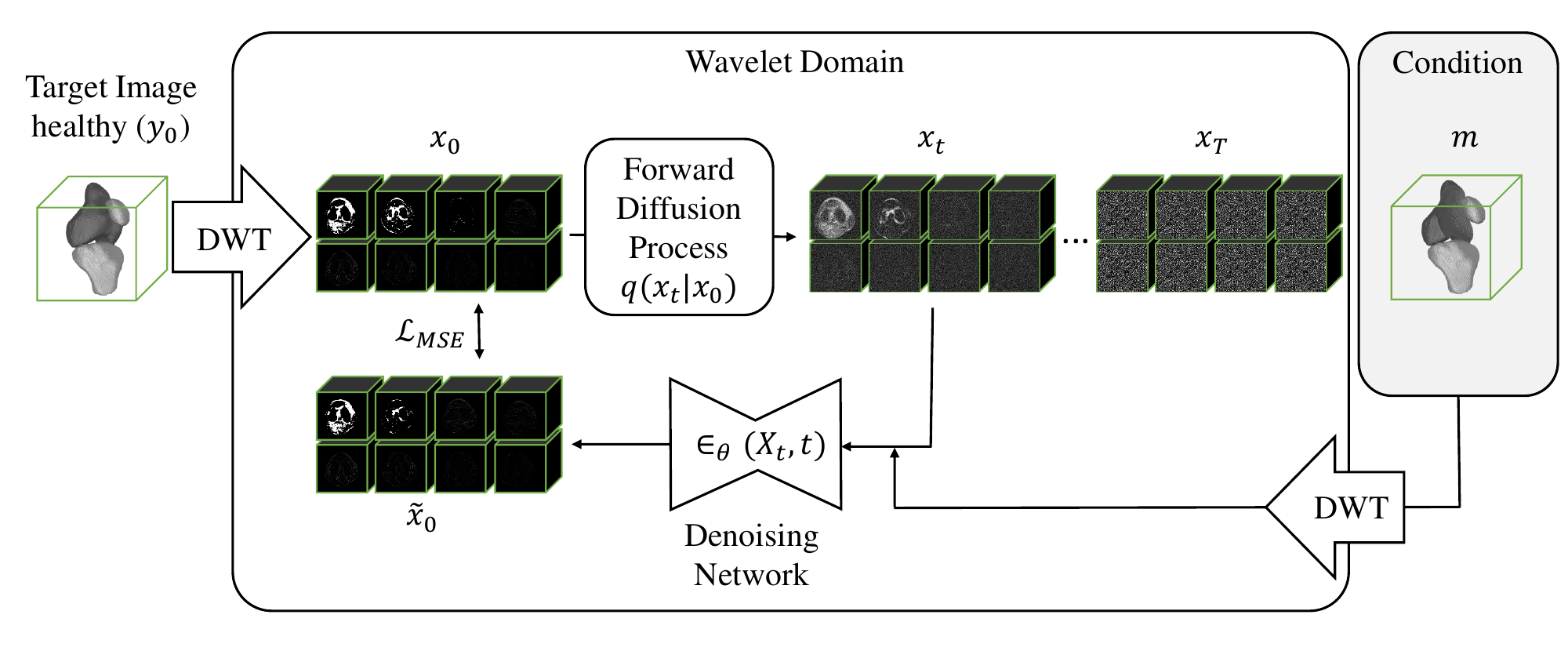}
    \caption{During training, the network is tasked with predicting the denoised image $\tilde{x}_0$ based on $X_t := (x_t, DWT(m))$ at a random time step $t$. The loss is then calculated between the prediction $\tilde{x}_0$ and $x_0$. Figure inspired by~\cite{friedrich2024cwdm}.\label{fig:Fig_Method_Inp}}
\end{figure}

Unlike~\cite{durrer2024denoising,wehrli2025generating}, who inpaint image intensities, we first segment the volume (as explained in Section~\ref{subsec:SEG}) and then do inpainting in the segmented space. We argue that this simplifies the enforcement of anatomical priors, since the final objective is to obtain a target morphology. Before using the Wavelet Diffusion Model (WDM) by~\cite{friedrich2024wdm} we first mask the trochlear region (with a 30~$mm$ offset around the patella), as shown in \autoref{fig:Fig_Method} between Step~2 and Step~3, resulting in our condition $m$ used to train the WDM, similar to~\cite{friedrich2024cwdm}. \autoref{fig:Fig_Method_Inp} shows our adapted version of the WDM for inpainting: The wavelet coefficients $x_0$ are obtained through a DWT of the ground truth $y_0$. Noise is added to $x_0$, resulting in $x_t = \sqrt{\bar{\alpha}_t}x_0 + \sqrt{1 - \bar{\alpha}_t} \epsilon, \quad \text{where} \quad \epsilon \sim \mathcal{N}(0, \mathbf{I})$ with $\alpha_t = 1 - \beta_t$ and $\bar{\alpha}_t = \prod_{s=1}^{t} \alpha_s$. Concatenating $x_t$ and the wavelet transformed masked image $m$ results in $ X_t := (x_t, DWT(m)) $. During training, $t$ is randomly chosen at each iteration. $X_t$ and $t$ are the inputs to the neural network $\epsilon_{\theta}$, which predicts the denoised $\tilde{x}_0$.
The model is trained using the MSE loss between $\tilde{x}_0$ and $x_0$.
During inference, the model is iteratively evaluated for $t = T, ..., 1$ with  $T$ = 1000 . Afterwards, the final prediction is obtained through an IDWT of $\tilde{x}_0$.

All used images have a resolution of $256^3$. The network was trained with batch size 1 and learning rate $1 \times 10^{-5}$ until convergence (50'000 iterations). T was set to 1000. More implementation details can be found at \url{https://wehrlimi.github.io/sr-3d-planning/}. For visualization purposes we use Marching Cubes to convert the inpainted segmentation to a mesh~\cite{lorensen1998marching}.

\section{Experiments}
\label{sec:experiments}
\subsubsection{Data}
\label{sec:data}
We evaluated our pipeline using two datasets: The publicly available fastMRI dataset (\url{https://fastmri.med.nyu.edu/})~\cite{knoll2020fastmri,zbontar2018fastmri} and an in-house dataset for patients diagnosed with TD. The entire pipeline, up to Step~3 (see \autoref{fig:Fig_Method}), is applied identically to both datasets.
For Step~3, we exclusively used data from fastMRI to train. We considered only subjects that have an axial proton density fat-saturated sequence and have all 3 volumes. Resulting in 1'568 subjects. We assume that this dataset represents a healthy trochlear shape representation. We split it in 80\% (1'254) for training, 20\% for testing of the model. 
Additionally, we used our in-house dataset with TD patients to evaluate our workflow on pathological data. It was anonymized and granted an exemption by the Ethikkommission Nordwest- und Zentralschweiz (Req-2024-01188). The described methods were carried out according to the relevant guidelines and regulations. From the available 52 patients, we excluded 27 patients having a subluxated patella (during the MR scan) or intense swelling (20) or no registered volumes (7) - Step~1 does not work then. The trained WDM was applied for inference to 25 TD patients (26 scans) to generate pseudo-healthy plans, offering insight into how their trochlear region might appear in the absence of pathology. The final image in \autoref{fig:Fig_Method} illustrates a 3D anomaly map, clearly indicating by bright colors that the patient's trochlea is abnormally shallow.

\subsubsection{Medical Evaluation}
We compared 26 knee MR scans of 25 patients with TD before and after inpainting using sulcus angle (SA), trochlear groove depth (TGD) and Déjour classification. The measurements were performed by a deputy attending orthopedic surgeon using 3D Slicer~\cite{slicer,kikinis20133d} Version 5.6.2.

\subsubsection{Image Quality evaluation}
Due to the lack of ground-truth high-resolution knee MR scans, we were not able to quantitatively verify Step~1 on our data. Therefore, we evaluated it qualitatively. The method was adapted from~\cite{mcginnis2023single}, who stated that they surpass State-of-the-Art (SOTA) methods in super-resolution reconstruction. To evaluate Step~2, we used the Dice Score. The super-resolved subset (see Section~\ref{subsec:SEG}) was randomly split into 90\% training (36 volumes) and 10\% test (4 volumes). To evaluate Step~3, we masked out a healthy knee region in the fastMRI test set, inpainted it using our proposed approach, and measured the deviation from the, in this case known, ground truth morphology using MSE.

\section{Results}
\subsection{Medical Evaluation}
\begin{figure}[h]
    \centering
    \includegraphics[width=0.9\linewidth]{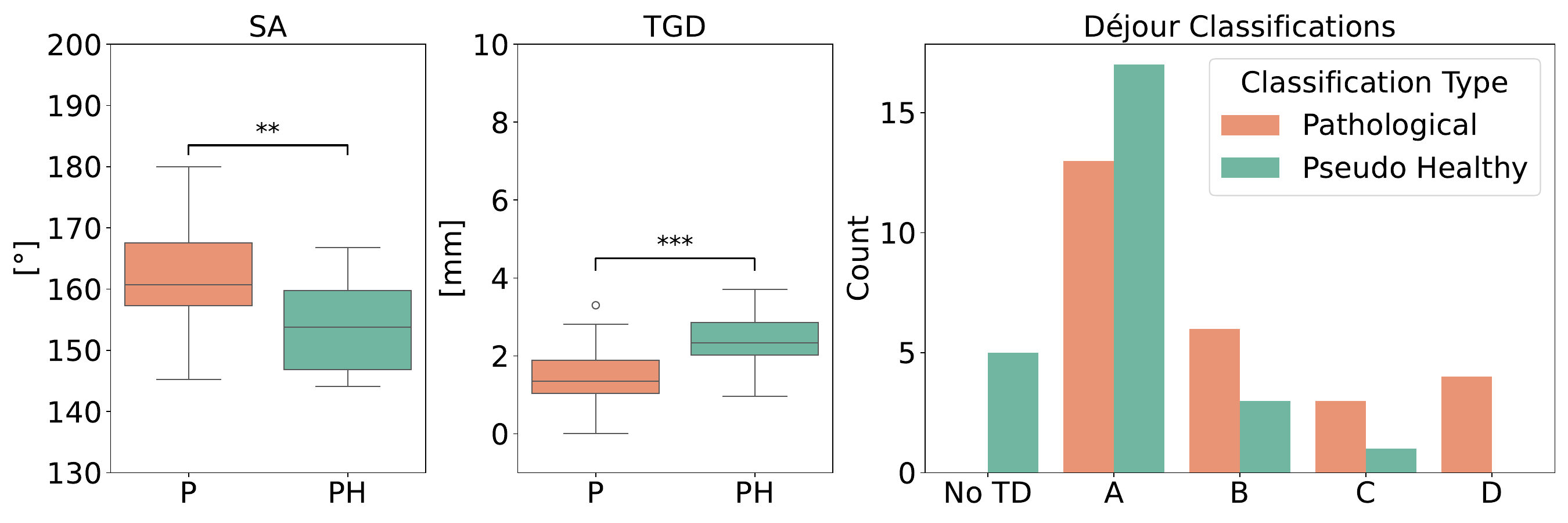}
    \caption{Left: The difference between the SA in the 22 scans is significant. Middle: Significant difference in TGD. Right: The pseudo-healthy versions show an evident reduction in severity according to the Déjour criterion.}
    \label{fig:Medical_Evaluation}
\end{figure}

\begin{figure}[h]
    \centering
    \includegraphics[width=0.95\linewidth]{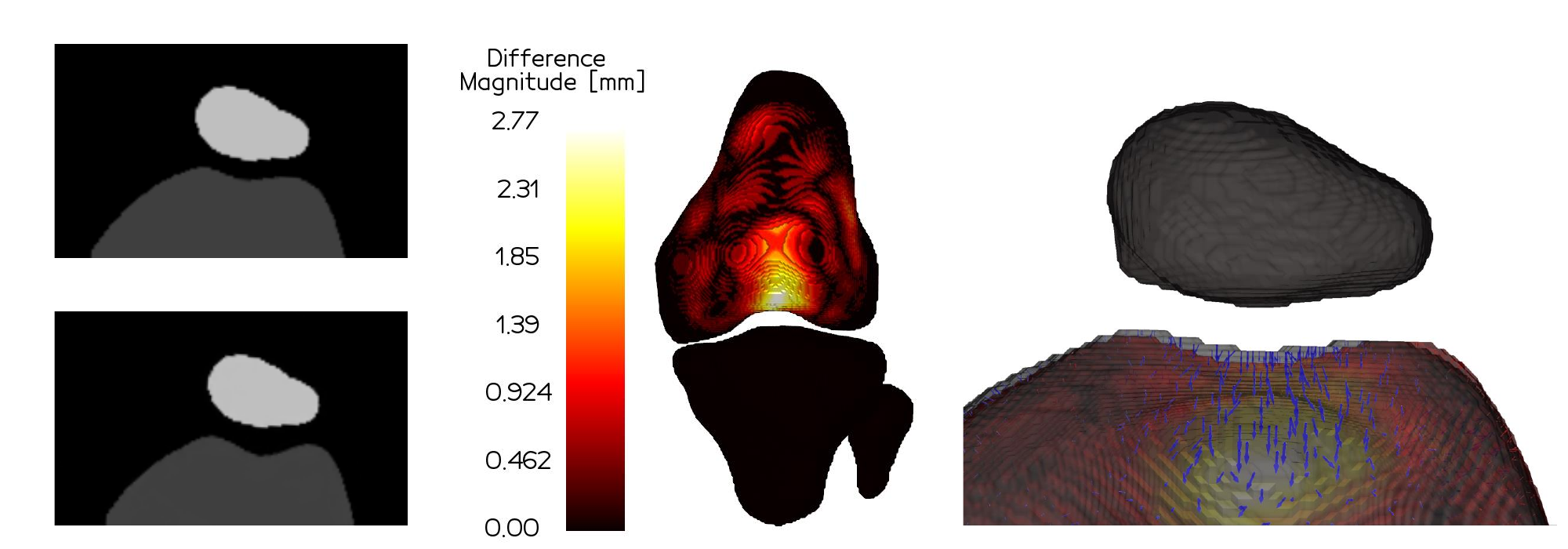}
    \caption{Left (top): Segmentation before inpainting of a patient with TD. Left (bottom): After inpainting. Middle: Difference Magnitude in 3D of the same patient, visualized w.o. patella. Right: Axial view that shows the difference between the original and morphological target. Blue arrows indicate where bone has to be removed. Red arrows indicate where bone could be added.}
    \label{fig:Mesh_result_qualitative}
\end{figure}

Of 26 pathological scans, 14 showed a clear reduction in the severity of TD in the pseudo-healthy versions according to the Déjour criterion by one or more stages. \autoref{fig:Medical_Evaluation} (right) demonstrates this shift. None of the samples were rated more severe than before.
In 22 patients the SA and TGD could be measured before (due to the nature of TD) and after inpainting. Regarding SA, a significant shift to a less shallow trochlea (from a mean of 162 degrees to a mean of 154 degrees) was observed (p value = 0.0011) by the Wilcoxon signed-rank test. The increase of the TGD is significant as well (p = 0.00006, from a mean of 1.48~$mm$ to 2.33~$mm$). A qualitative example is visualized in \autoref{fig:Mesh_result_qualitative}.

\subsection{Image Quality evaluation}
\subsubsection{Step 1: INR}
\begin{figure}[h]
    \centering
    \includegraphics[width=0.75\linewidth]{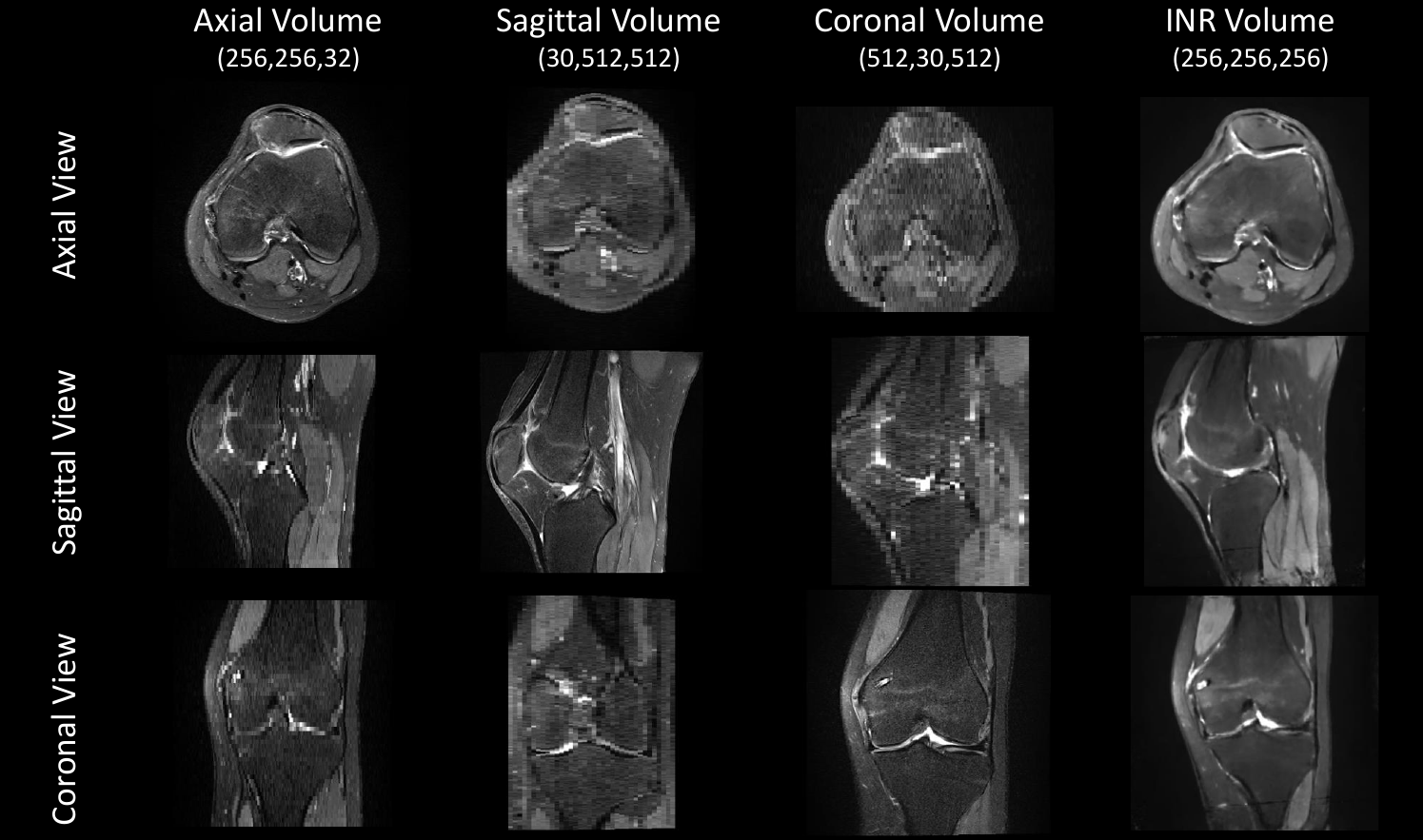}
    \caption{Left to right: Axial volume, Sagittal volume, Coronal volume and on the very right the INR volume, combining the information of all others.}
    \label{fig:result_INR_qualitative}
\end{figure}

Qualitative results for a patient with TD can be seen in \autoref{fig:result_INR_qualitative}. The very right column shows the INR volume, that has high-resolution images in all 3 views.

\subsubsection{Step 2: Segmentation}
We compare our custom-trained segmentation model (MonaiLabel) against SOTA bone knee segmentation tools. Note that we are not claiming to have the best performing segmentation model architecture. We demonstrate, see \autoref{table:segmentation}, that with our manual segmentations one can achieve superior femur segmentation in super-resolved knee MR scans - a crucial step to create pseudo-healthy trochleas. Totalsegmentator~\cite{d2025totalsegmentator} has no module to segment the full knee MR scan at once. We used the \texttt{appendicular\_bones\_mr} module to segment tibia, fibula and patella and \texttt{total\_mr} for the femur. SegmentAnyBone~\cite{gu2025segmentanybone} is only able to perform binary segmentation.

\begin{table}[h]
\centering
\caption{Quantitative segmentation results (Dice score ± std). \label{table:segmentation}}
\begin{tabular}{lccc}
& \textbf{MonaiLabel (Ours)} & \textbf{SegmentAnyBone} & \textbf{TotalSegmentator} \\
Femur     & \textbf{0.9402 ± 0.0187} & n.a. & 0.8970 ± 0.0351 \\
Tibia     & \textbf{0.9266 ± 0.0089} & n.a. & 0.8869 ± 0.0333 \\
Patella   & 0.7870 ± 0.1087 & n.a. & \textbf{0.8104 ± 0.0593} \\
Fibula    & \textbf{0.8784 ± 0.0699} & n.a. & 0.7232 ± 0.0672 \\
Mean      & \textbf{0.8830 ± 0.0516} & n.a. & 0.8294 ± 0.0487 \\  
Binary    & \textbf{0.9262 ± 0.0096} & 0.7392 ± 0.1276 & 0.8807 ± 0.0181 \\
\end{tabular}
\end{table} 

\subsubsection{Step 3: Wavelet Diffusion Model}
On the fastMRI test set, we observed an MSE of: $0.0017~\pm{0.0015}$.

\section{Conclusion}
We propose a pipeline for the generation of surgical target morphologies for trochleoplasty from conventional clinical MR scans. The pipeline has the potential to be used intraoperatively for surgical navigation. While this work focused on TD, the pipeline is not limited to this pathology and could be adapted to other surgical applications. However, our method also comes with certain limitations: A sufficiently large dataset representing healthy anatomy is required to train the WDM. Further, the fastMRI dataset may not capture the ideal trochlear shape, could be age-biased, and is likely skewed toward a North American population. Additionally, conventional MR scans must be registered to be used. Also, scans with a dislocated patella are not suitable for our current approach. Segmentation quality remains a bottleneck, and could be improved with more extensive manual annotations. In conclusion, we propose a pipeline that generates super-resolved, patient-specific 3D pseudo-healthy target morphologies from conventional MR scans. It paves the way for more reproducible, minimally invasive and patient-tailored surgery procedures. The next major step is to validate this pipeline intraoperatively using a navigation system, moving toward integration into real-world surgical workflows.

\begin{credits}
\subsubsection{\ackname} This work was financially supported by the Werner Siemens Foundation through the MIRACLE II project.

\subsubsection{\discintname}
The authors have no competing interests to declare that are relevant to the content of this article.
\end{credits}

\bibliographystyle{splncs04} 
\bibliography{Paper-0001.bib}

\begin{thebibliography}{10}
\providecommand{\url}[1]{\texttt{#1}}
\providecommand{\urlprefix}{URL }
\providecommand{\doi}[1]{https://doi.org/#1}

\bibitem{slicer}
{3D Slicer Community}: 3d slicer - a free, open source, and extensible image computing platform. \url{https://www.slicer.org/} (2025), accessed: 2025-07-03

\bibitem{batailler2018trochlear}
Batailler, C., Neyret, P.: Trochlear dysplasia: imaging and treatment options. EFORT open reviews  \textbf{3}(5),  240--247 (2018)

\bibitem{beaufils2012trochleoplasty}
Beaufils, P., Thaunat, M., Pujol, N., Scheffler, S., Rossi, R., Carmont, M.: Trochleoplasty in major trochlear dysplasia: current concepts. Sports Medicine, Arthroscopy, Rehabilitation, Therapy \& Technology  \textbf{4}, ~1--8 (2012)

\bibitem{blond2015arthroscopic}
Bl{\o}nd, L.: Arthroscopic deepening trochleoplasty: the technique. Operative Techniques in Sports Medicine  \textbf{23}(2),  136--142 (2015)

\bibitem{blond2023trochlear}
Bl{\o}nd, L., Barfod, K.W.: Trochlear shape and patient-reported outcomes after arthroscopic deepening trochleoplasty and medial patellofemoral ligament reconstruction: a retrospective cohort study including mri assessments of the trochlear groove. Orthopaedic Journal of Sports Medicine  \textbf{11}(5),  23259671231171378 (2023)

\bibitem{cerveri2018stacked}
Cerveri, P., Belfatto, A., Baroni, G., Manzotti, A.: Stacked sparse autoencoder networks and statistical shape models for automatic staging of distal femur trochlear dysplasia. The International Journal of Medical Robotics and Computer Assisted Surgery  \textbf{14}(6),  e1947 (2018)

\bibitem{cerveri2019representative}
Cerveri, P., Belfatto, A., Manzotti, A.: Representative 3d shape of the distal femur, modes of variation and relationship with abnormality of the trochlear region. Journal of biomechanics  \textbf{94},  67--74 (2019)

\bibitem{d2025totalsegmentator}
D'Antonoli, T.A., Berger, L.K., Indrakanti, A.K., Vishwanathan, N., Weiss, J., Jung, M., Berkarda, Z., Rau, A., Reisert, M., K{\"u}stner, T., et~al.: Totalsegmentator mri: Robust sequence-independent segmentation of multiple anatomic structures in mri. Radiology  \textbf{314}(2),  e241613 (2025)

\bibitem{diaz2024monai}
Diaz-Pinto, A., Alle, S., Nath, V., Tang, Y., Ihsani, A., Asad, M., P{\'e}rez-Garc{\'\i}a, F., Mehta, P., Li, W., Flores, M., et~al.: Monai label: A framework for ai-assisted interactive labeling of 3d medical images. Medical Image Analysis  \textbf{95},  103207 (2024)

\bibitem{durrer2024denoising}
Durrer, A., Wolleb, J., Bieder, F., Friedrich, P., Melie-Garcia, L., Ocampo~Pineda, M.A., Bercea, C.I., Hamamci, I.E., Wiestler, B., Piraud, M., et~al.: Denoising diffusion models for 3d healthy brain tissue inpainting. In: MICCAI Workshop on Deep Generative Models. pp. 87--97. Springer (2024)

\bibitem{fang2024patient}
Fang, X., Deng, H.H., Kuang, T., Xu, X., Lee, J., Gateno, J., Yan, P.: Patient-specific reference model estimation for orthognathic surgical planning. International Journal of Computer Assisted Radiology and Surgery  \textbf{19}(7),  1439--1447 (2024)

\bibitem{friedrich2024cwdm}
Friedrich, P., Durrer, A., Wolleb, J., Cattin, P.C.: cwdm: Conditional wavelet diffusion models for cross-modality 3d medical image synthesis. arXiv preprint arXiv:2411.17203  (2024)

\bibitem{friedrich2024wdm}
Friedrich, P., Wolleb, J., Bieder, F., Durrer, A., Cattin, P.C.: Wdm: 3d wavelet diffusion models for high-resolution medical image synthesis. In: MICCAI Workshop on Deep Generative Models. pp. 11--21. Springer (2024)

\bibitem{gu2025segmentanybone}
Gu, H., Colglazier, R., Dong, H., Zhang, J., Chen, Y., Yildiz, Z., Chen, Y., Li, L., Yang, J., Willhite, J., et~al.: Segmentanybone: A universal model that segments any bone at any location on mri. Medical Image Analysis p. 103469 (2025)

\bibitem{hasler2016patella}
Hasler, C.C., Studer, D.: Patella instability in children and adolescents. EFORT open reviews  \textbf{1}(5),  160--166 (2016)

\bibitem{kikinis20133d}
Kikinis, R., Pieper, S.D., Vosburgh, K.G.: 3d slicer: a platform for subject-specific image analysis, visualization, and clinical support. In: Intraoperative imaging and image-guided therapy, pp. 277--289. Springer (2013)

\bibitem{knoll2020fastmri}
Knoll, F., Zbontar, J., Sriram, A., Muckley, M.J., Bruno, M., Defazio, A., Parente, M., Geras, K.J., Katsnelson, J., Chandarana, H., et~al.: fastmri: A publicly available raw k-space and dicom dataset of knee images for accelerated mr image reconstruction using machine learning. Radiology: Artificial Intelligence  \textbf{2}(1),  e190007 (2020)

\bibitem{lee2024shallow}
Lee, J.Y., Kim, S.E., Kwon, O.H., Kim, Y., Son, T.g., Han, H.S., Ro, D.H.: Shallow trochlear groove and narrow medial trochlear width at the proximal trochlea in patients with trochlear dysplasia: A three-dimensional computed tomography analysis. Knee Surgery, Sports Traumatology, Arthroscopy  \textbf{32}(6),  1434--1445 (2024)

\bibitem{lorensen1998marching}
Lorensen, W.E., Cline, H.E.: Marching cubes: A high resolution 3d surface construction algorithm. In: Seminal graphics: pioneering efforts that shaped the field, pp. 347--353 (1998)

\bibitem{mcginnis2023single}
McGinnis, J., Shit, S., Li, H.B., Sideri-Lampretsa, V., Graf, R., Dannecker, M., Pan, J., Stolt-Ans{\'o}, N., M{\"u}hlau, M., Kirschke, J.S., et~al.: Single-subject multi-contrast mri super-resolution via implicit neural representations. In: International Conference on Medical Image Computing and Computer-Assisted Intervention. pp. 173--183. Springer (2023)

\bibitem{myronenko20183d}
Myronenko, A.: 3d mri brain tumor segmentation using autoencoder regularization. In: International MICCAI brainlesion workshop. pp. 311--320. Springer (2018)

\bibitem{nacey2017magnetic}
Nacey, N.C., Geeslin, M.G., Miller, G.W., Pierce, J.L.: Magnetic resonance imaging of the knee: an overview and update of conventional and state of the art imaging. Journal of Magnetic Resonance Imaging  \textbf{45}(5),  1257--1275 (2017)

\bibitem{saragadam2023wire}
Saragadam, V., LeJeune, D., Tan, J., Balakrishnan, G., Veeraraghavan, A., Baraniuk, R.G.: Wire: Wavelet implicit neural representations. In: Proceedings of the IEEE/CVF Conference on Computer Vision and Pattern Recognition. pp. 18507--18516 (2023)

\bibitem{van2014statistical}
Van~Haver, A., Mahieu, P., Claessens, T., Li, H., Pattyn, C., Verdonk, P., Audenaert, E.: A statistical shape model of trochlear dysplasia of the knee. The Knee  \textbf{21}(2),  518--523 (2014)

\bibitem{wehrli2025generating}
Wehrli, M., Durrer, A., Friedrich, P., Buchakchiyskiy, V., Mumme, M., Li, E., Lehoczky, G., Hasler, C.C., Cattin, P.C.: Generating 3d pseudo-healthy knee mr images to support trochleoplasty planning. International Journal of Computer Assisted Radiology and Surgery pp.~1--8 (2025)

\bibitem{zbontar2018fastmri}
Zbontar, J., Knoll, F., Sriram, A., Murrell, T., Huang, Z., Muckley, M.J., Defazio, A., Stern, R., Johnson, P., Bruno, M., et~al.: fastmri: An open dataset and benchmarks for accelerated mri. arXiv preprint arXiv:1811.08839  (2018)

\end{thebibliography}

\end{document}